\documentclass[10pt,twocolumn,letterpaper]{article}
\usepackage{cvpr}
\usepackage{times}
\usepackage{epsfig}
\usepackage{graphicx}
\usepackage{amsmath}
\usepackage{amssymb}
\usepackage{booktabs} 
\usepackage{comment}
\usepackage{wrapfig}
\usepackage{multirow}
\usepackage[table,xcdraw]{xcolor}
\usepackage{graphicx} 

\usepackage{enumitem}
\usepackage{threeparttable}
\usepackage[accsupp]{axessibility} 

\usepackage[none]{hyphenat}
\usepackage{enumitem}

\usepackage{amsmath}
\usepackage{ragged2e}
\usepackage{amsmath}
\usepackage{pifont}

\usepackage[noadjust]{cite} 

\usepackage{hyperref}

\begin{document}

\title{Keystroke Dynamics Against Academic Dishonesty in the Age of LLMs}

\author{
    Debnath Kundu\textsuperscript{$\dagger$, 1}, 
    Atharva Mehta\textsuperscript{$\dagger$, 2}, 
    Rajesh Kumar\textsuperscript{$\ddagger$, 3} \\
    Naman Lal\textsuperscript{$\dagger$, 4},
    Avinash Anand\textsuperscript{$\dagger$, 5},
    Apoorv Singh\textsuperscript{$\dagger$, 6},
    Rajiv Ratn Shah\textsuperscript{$\dagger$, 7}
    \\
    \textsuperscript{$\dagger$}MIDAS Lab IIIT Delhi India, 
    \textsuperscript{$\ddagger$}Bucknell University USA \\
    \normalsize\{\textsuperscript{1}\texttt{debnath22026}, 
    \textsuperscript{2}\texttt{atharva20038\}@iiitd.ac.in},
    \textsuperscript{3}\texttt{rajesh.kumar@bucknell.edu}\\
    \normalsize\textsuperscript{4}\texttt{namanlal.lal@gmail.com}, 
    \{\textsuperscript{5}\texttt{avinasha,}
    \textsuperscript{6}\texttt{apoorv17027,}
    \textsuperscript{7}\texttt{rajivratn\}@iiitd.ac.in}
}

\maketitle
\thispagestyle{empty}

\begin{abstract}
The transition to online examinations and assignments raises significant concerns about academic integrity. Traditional plagiarism detection systems often struggle to identify instances of intelligent cheating, particularly when students utilize advanced generative AI tools to craft their responses. This study proposes a keystroke dynamics-based method to differentiate between bona fide and assisted writing within academic contexts. To facilitate this, a dataset was developed to capture the keystroke patterns of individuals engaged in writing tasks, both with and without the assistance of generative AI. The detector, trained using a modified TypeNet architecture, achieved accuracies ranging from $74.98\%$ to $85.72\%$ in condition-specific scenarios and from $52.24\%$ to $80.54\%$ in condition-agnostic scenarios. The findings highlight significant differences in keystroke dynamics between genuine and assisted writing. The outcomes of this study enhance our understanding of how users interact with generative AI and have implications for improving the reliability of digital educational platforms.
\end{abstract}

\small \thanks{The IEEE International Joint Conference on Biometrics
979-8-3503-9494-8/24/\$31.00 \copyright 2024 IEEE.}

\section{Introduction}
With the rapid advancements in generative AI technology and increased reliance on virtual educational environments, cheating in online examinations has significantly escalated. Many of us can now confidently identify AI-generated plagiarism, yet it remains a challenging task, especially in large class settings. Testing agencies have implemented several online proctoring measures to uphold integrity during examinations. These include prohibiting tab changes, disabling the copy-paste function, and preventing the minimization of the exam window. Such restrictions aim to mimic the controlled environment of in-person exams and ensure a level playing field. However, these measures fall short of addressing all forms of academic dishonesty. Traditional plagiarism detection tools like \textit{Turnitin} and \textit{Urkund} work by comparing submissions against existing sources. While these tools effectively detect direct plagiarism, they often fail to recognize more sophisticated cheating methods, such as extensive paraphrasing or external assistance via generative AI.

\begin{figure}[htp]
    \centering
    \includegraphics[width=2.55in, height = 1.3in]{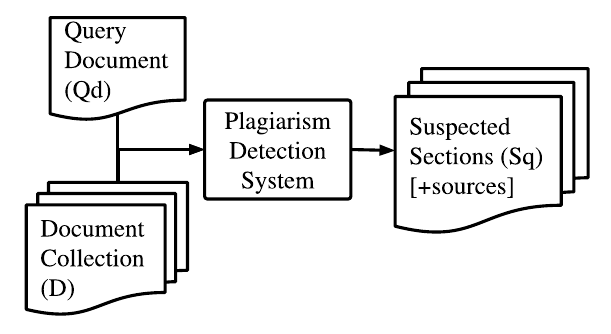}
    \caption{The framework of existing plagiarism detectors such as Turnitin and Urkund, which use content-based word-for-word similarities \cite{alzahrani2011understanding, rao2008plagiarism}.}
    \label{fig:plag_detection_tools}
\end{figure}

Figure \ref{fig:plag_detection_tools} illustrates a framework diagram for existing systems that process a primary input, typically a query or suspicious document ($Q_d$), and an optional reference collection, such as the web ($D$). These systems output identified suspicious fragments and their sources of plagiarism ($S_q$) when available. However, they are ineffective against more sophisticated forms of cheating, including text manipulation, extensive paraphrasing, and assistance from another individual \cite{rao2008plagiarism}. The emergence of advanced large language models like ChatGPT and Gemini has further complicated the landscape. These AI systems can generate high-quality, human-like essays within seconds, posing new challenges for current plagiarism detection mechanisms. The capability of these tools to produce tailored, undetectable responses is particularly problematic for traditional plagiarism detection software, which relies on direct text comparison.

Numerous studies have explored plagiarism within academic circles, outlining various types of plagiarism and potential detection methods \cite{anderson2009avoiding, pecorari2008academic, rao2008plagiarism, roig2015avoiding}. Linguistic analyses have indicated that identifying specific contexts is crucial for effective plagiarism detection \cite{hoq2024detecting, zhang2011coarse}. For instance, in introductory computing courses, cheating involving ChatGPT can be detected by comparing student code submissions with those generated by ChatGPT, achieving an accuracy of 90\% or better. Notable differences often include conciseness, reduced variable duplication, simpler control flow, and a preference for ternary operators in the code produced by ChatGPT \cite{hoq2024detecting}.

The ability to distinguish between AI-generated and human-generated text has garnered significant interest from a wide range of researchers since the introduction of ChatGPT \cite{pu2023chatgpt, dugan2023real, Herbold2023, CASAL2023100068, BAiVsHuman, AiVsHumanCS1, AiVsHuman}. While traditional natural language processing techniques concentrated on the content of responses, recent studies show that behavioral signals, from the text's production (such as typing behavior) to its consumption (including reading or gaze behavior), offer valuable insights about the text \cite{khurana2023synthesizing, BehavioralNLP2}. Typing behavior captured by keystroke dynamics provides extensive information about text production processes. This method has become a well-established, non-intrusive behavioral modality for user authentication, which we are adapting to detect plagiarism \cite{MONROSE2000351, Joyce1990, 62613, press1980authentication, teh2013survey, TypingPhoneBTAS2016}.

The Educational Testing Service (ETS) has effectively utilized keystroke log features to predict essay scores. Models specific to tasks, genres, and contexts, trained on features such as time spent on tasks, burst lengths in words and pauses within words, have demonstrated good generalizability across various prompts \cite{deane2015exploring}.

Another prevalent form of academic dishonesty in online or distance learning examinations is impersonation, often referred to as contract cheating \cite{agarwal-nancy-contract}. Keystroke analysis can address this issue by comparing the keystroke dynamics of the registered examinee with those submitted during the examination\cite{kochegurova2022hidden, mungai2017using}.

\begin{figure}[ht]
    \centering
    \includegraphics[width=3in, height=0.8in]{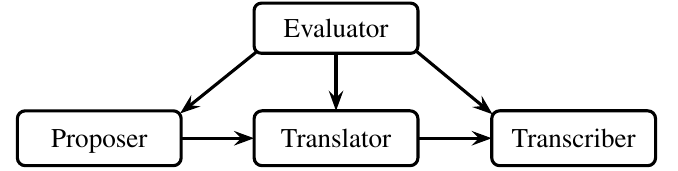}
    \caption{The cognitive model of the writing process includes four stages: \textit{Proposer}, where ideas are generated and tasks prepared, marked by initial pauses and pauses at sentence boundaries; \textit{Translator}, involving the fluent conversion of ideas into language, measured by uninterrupted text production sequences; \textit{Transcriber}, focusing on orthographic proficiency and motor skills, evident in pauses within words and immediate spelling corrections; and \textit{Evaluator}, which entails editing and reviewing, identifiable by jumps to different text parts for extensive edits \cite{zhang2021using}.}
    \label{fig:hayes}
\end{figure}

Previous research in writing cognition, including Hayes and Flower’s model \cite{hayes2012modeling}, identified key sub-processes in writing: idea generation, translation of ideas into text, typing, and revising. Figure \ref{fig:hayes} depicts a simplified version of Hayes’ cognitive writing model \cite{flower1981cognitive}, which includes four sub-processes: \textit{Proposer}, \textit{Translator}, \textit{Transcriber}, and \textit{Evaluator}. These models suggest that authentic writing tasks require more planning and revising than assisted writing, highlighting the importance of these processes in effective writing. Inspired by previous studies \cite{balagani2013investigating, zhang2021using, flower1981cognitive, lim2015using}, we hypothesize and investigate whether keystroke dynamics and linguistic analyses can capture the distinctive cognitive processes behind bona fide and assisted writing.

In other words, the cognitive stages of the writing process are closely linked to distinctive keystroke dynamics, providing a window into the writer's cognitive states. For example, the \textit{Proposer} stage could be characterized by initial pauses and pauses at sentence boundaries, reflecting the cognitive load of idea generation. In the \textit{Translator} stage, fluent typing with minimal interruptions indicates efficient translation of thoughts to text. The \textit{Transcriber} stage is characterized by pauses within words and immediate corrections, highlighting the focus on spelling and grammatical accuracy. Finally, the \textit{Evaluator} stage involves extensive cursor movements and edits across the text, indicative of thorough reviewing and editing processes. We can infer whether a text was produced by following a natural writing process or plagiarized by analyzing keystroke patterns—like pause lengths, typing rhythm, and editing behaviors.

In essence, our goal is to train and evaluate a suite of classifiers—denoted as---\( f_K \)for keystroke dynamics \( X_K \) to detect academic dishonesty by classifying text and corresponding keystrokes submitted by students as bona fide (\( y=0 \)) or AI-assisted (\( y=1 \)), across various academic tasks and contexts. The evaluation is conducted under diverse training and testing setups to explore the impact of task complexity and contextual variations on classifier performance. In summary, our main contributions are as follows:

\begin{itemize}
\itemsep-0.15em
\item The study introduces a keystroke dynamics-based method for detecting academic dishonesty, employing the TypeNet model to differentiate between bona fide and AI-assisted writing.
\item It proposes a unique dataset capturing keystroke patterns during various writing tasks for training and evaluating the plagiarism detection models.
\item Conducts a comprehensive evaluation that reveals the method's effectiveness under different scenarios, including variations across keyboards, users, and contexts, while also discussing the inherent challenges and limitations in typing behavior analysis.
\item The paper highlights the potential of keystroke dynamics as a supplementary tool to traditional plagiarism detection methods and proposes future research directions to enhance algorithm accuracy and ensure fairness in academic assessments.
\end{itemize}

The remainder of the paper is organized as follows: Section \ref{RelatedWorks} discusses related work; Section \ref{ExperimentalSetup} outlines the experimental design; Section \ref{ResultsDiscussion} presents the results and discussion; Section \ref{limitations} enumerates the limitations; and Section \ref{ConclusionFutureWorks} concludes the study, offering possible future directions.
 
\section{Related work}
\label{RelatedWorks}
The persistent challenge of academic dishonesty, specifically plagiarism, has spurred numerous technological advancements aimed at detection and prevention. Historically, tools like Turnitin have relied on textual analysis to detect similarities between submitted work and existing materials \cite{patel2011evaluation}. Despite these efforts, students have found ways to bypass such systems using sophisticated methods such as paraphrasing or employing ghostwriters. Consequently, recent studies have adopted machine learning approaches, notably autoencoders, and LSTM networks, to analyze patterns in student assessment data. Kamalov et al. \cite{kamalov2021machine} highlighted the efficacy of combining LSTMs with kernel density estimation for detecting discrepancies in student performance, which is crucial for identifying instances of academic dishonesty during post-exam analyses. Additionally, surveillance technologies, including video monitoring and IP address tracking, have enhanced the integrity of examinations \cite{roa2022automated, tiong2021cheating}.

Keystroke dynamics, which record neuro-physiological characteristics comparable to handwritten signatures, have been widely studied. In the domain of continuous user authentication, these dynamics are studied to assess latency times between keystrokes, thus enhancing password security and thwarting unauthorized access \cite{ayotte2020fast, monrose1999password, TypingPhoneBTAS2016}. The applications of keystroke dynamics also extend to uncovering the anonymity of online content, automating the estimation of soft biometrics \cite{KeystrokeSoftBiometrics}, detecting lies \cite{Monaro2018Covert}, identifying deceptive reviews \cite{banerjee2014_emnlp}, and discerning fake profiles on social media \cite{morales2020keystroke, banerjee2014_emnlp, kuruvilla2024spotting}. Additionally, keystroke analysis is crucial for authorship attribution and profiling, enabling the prediction of demographic characteristics such as age and gender based on typing patterns \cite{tsimperidis2018r, plank2018predicting}. Importantly, Goodman et al. \cite{email-author-goodman} have applied stylometric features with raw keystroke data to precisely predict the authors of emails, thereby confirming the accuracy of keystroke dynamics in authorship attribution.

The inherent capabilities of keystroke dynamics in authorship attribution and user authentication indicate their potential applicability for plagiarism detection. The accurate measurement of typing rhythms and patterns may reveal the authenticity of the text production process. For instance, abrupt changes in typing speed or key pressure could signify a transition from original writing to copied text, thereby serving as indicators of plagiarism. This method could significantly aid traditional content-based plagiarism detection systems because it is agnostic to language and content, concentrating instead on the behavioral aspects exhibited during writing.

While keystroke dynamics have demonstrated efficacy in diverse fields, their application in plagiarism detection remains underexplored. This study addresses this gap by employing keystroke dynamics to distinguish between authentic and AI-assisted or plagiarized texts. Focusing on typing behavior under various cognitive loads, this research broadens the application of keystroke dynamics beyond simple authentication and authorship attribution, transforming them into a valuable tool for upholding academic integrity. The proposed models are designed to detect subtle variations in typing patterns and linguistic styles, presenting a novel method for identifying plagiarism in academic environments.

\section{Design of experiment}
\label{ExperimentalSetup}
\begin{figure}[htp]
    \centering
    \includegraphics[width=3in, height= 1.65in]{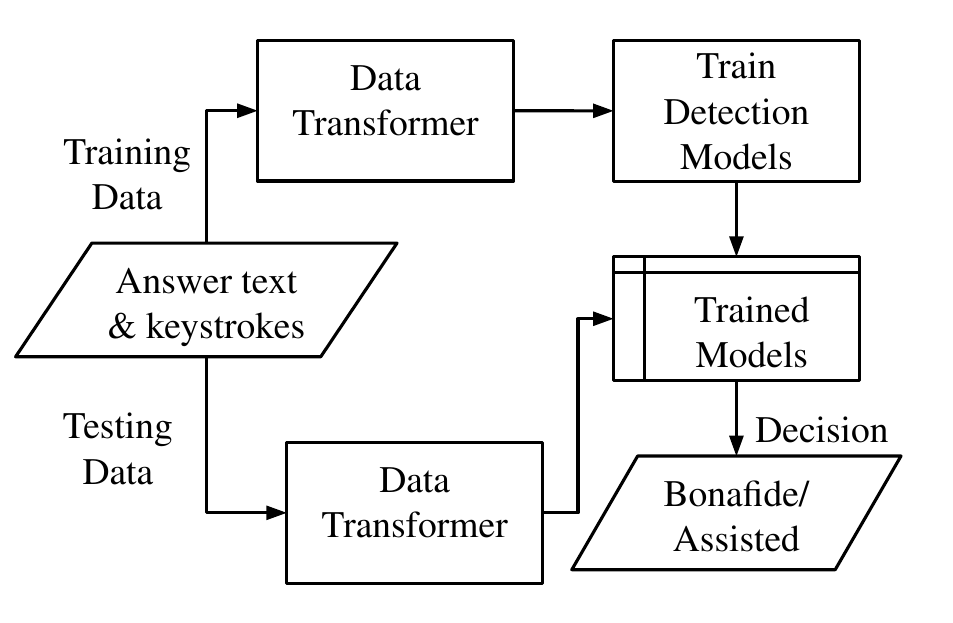}
    \caption{Illustrating the proposed detection framework that transforms training and testing data for use in training the detection models, which then contribute to deciding whether the output is bonafide or assisted.}
    \label{Proposedframework}
\end{figure}

The proposed framework, depicted in Figure \ref{Proposedframework}, outlines the experimental setup. Detailed descriptions of each component are provided in the following sections. The dataset and implementation details are made available via Github\footnote{https://github.com/ijcb-2024/keystroke-llm-plagiarism}. 

\subsection{Datasets} 
We utilize three distinct keystroke datasets, including two established collections from Stony Brook University (SBU) and SUNY Buffalo, alongside one custom dataset developed specifically for this study.

The SBU \cite{banerjee2014_emnlp} and Buffalo \cite{7823894} datasets were selected as they aligned with the experimental scenarios we intend to explore. Table \ref{tab:comparison} presents a comparative analysis of their data collection processes.

\textit{\textbf{The SBU dataset}} comprises both truthful and deceptive writings across two domains: restaurant business reviews and social issue essays on gun control and gay marriage. Participants were instructed to type freely and retype their entries, facilitating the analysis of natural and controlled typing behaviors, thus simulating bona fide and assisted typing conditions, respectively.

\textit{\textbf{The Buffalo dataset}} contains keystroke data from $157$ participants who transcribed fixed texts and responded to questions freely across three sessions using four different keyboards. This dataset is particularly valuable for examining how various keyboards affect typing behavior. This dataset's fixed text transcription component mimics assisted typing scenarios, where participants replicate given texts, resembling situations where content is mindlessly copied.

\begin{table}[htp]
\centering
\caption{Details of SBU, Buffalo, and the proposed datasets. }
\vspace{0.15in}
\begin{tabular}{|l|c|c|c|}
\hline
\textbf{Characteristics} & \textbf{SBU} & \textbf{Buffalo} & \textbf{Proposed} \\
\hline
Users & 196 & 148 & 40 \\
\hline
Free writing & \ding{51} & \ding{51} & \ding{51} \\
\hline
Fixed writing & \ding{51} & \ding{51} & \ding{51} \\
\hline
Keyboard labels & \ding{55} & \ding{51} & \ding{55} \\ \hline
Cognitive-information & \ding{55} & \ding{55} & \ding{51} \\
\hline
AI-assisted writing & \ding{55} & \ding{55} & \ding{51} \\
\hline
\end{tabular}
\label{tab:comparison}
\end{table}

\textbf{\textit{Proposed dataset}} 
While the SBU and Buffalo datasets align well with our research goals, they do not fully encompass the complexity of keystroke dynamics in controlled academic environments, particularly in scenarios involving bona fide and AI-assisted writing. An IRB-approved data collection protocol was established to replicate these scenarios in online examination settings.  

The proposed dataset comprised 40 students from various academic disciplines at a major, research-intensive university predominantly within STEM fields. This diversity ensured a comprehensive representation of typing skills and cognitive approaches. Each participant completed two separate sessions: In the first session, participants typed responses independently without any aids, showcasing their intrinsic typing abilities and cognitive processes. The second session saw the same participants using internet access and AI tools like ChatGPT to mimic conditions of potential academic misconduct. The questions posed were designed to elicit different levels of cognitive load, covering both general (opinion-based) and science-specific (fact-based) subjects that required analytical and methodological reasoning. Responses that did not meet established quality standards, such as a minimum word count, were omitted to preserve the integrity of the dataset.

Prior research has demonstrated that users exhibit distinct writing patterns when engaging with familiar versus unfamiliar topics \cite{lettner2010questions}. Specifically, writers take longer pauses to reflect on their text and organize their thoughts when confronted with less familiar topics, often using the backspace key to enhance textual coherence. In contrast, writing on familiar topics generally proceeds more rapidly and with greater consistency \cite{trezise2019detecting}. Accordingly, our study was designed to include two targeted sections, \textit{Opinion-based} and the \textit{Fact-based}, to explore these variations in writing behavior.

The \textit{Opinion-based} section comprised two open-ended essay questions addressing broad topics in society, environment, and current affairs. Participants were tasked with arguing for or against a specific proposition, such as, \textit{'Is AI a promising tool that can revolutionize industries and improve human life, or does it carry significant ethical risks, including job displacement and loss of control?'}

Meanwhile, the \textit{Fact-based} section included two scenario-driven questions centered on high-school science concepts. Participants were required to elaborate on their observations and rationalize the underlying scientific principles. For instance, one question posed was, \textit{'You are given a mixture of sand and salt. Design an experiment to separate the two components using appropriate laboratory equipment and techniques. Detail your step-by-step procedure and the principles underlying the separation process.'}

Participants crafted responses to four questions each session, structured to elicit one of six cognitive load levels defined by Balagani et al. \cite{balagani2013investigating}. Each typing session, capturing detailed keystroke dynamics such as key-down and key-up events, spanned 30 to 45 minutes. Participants were required to compose answers with a minimum of 300 characters per question and were permitted to refine their responses using the Delete or Backspace keys. This data, recorded by high-precision keystroke sensors accurate to ±200 microseconds, constitutes the Proposed dataset further detailed in our study.

Employing three distinct datasets facilitated a comprehensive analysis across diverse typing scenarios—free, fixed, bona fide, and assisted—thereby enabling a nuanced understanding of keystroke dynamics within academic settings. The targeted methodology of the custom dataset creation, in conjunction with the comparative analysis against the SBU and Buffalo datasets, enhances this understanding and contributes to the integrity of educational assessments.
 
\subsection{Training detection models}

To analyze keystroke dynamics, we implemented a deep learning-based approach utilizing an LSTM architecture, specifically the widely recognized LSTM model, TypeNet.

\textit{\textbf{TypeNet}} \cite{acien2021typenet} is a Long Short-Term Memory network developed to authenticate users by analyzing their typing patterns, including metrics such as typing speed, rhythm, and key press duration. The LSTM component is essential in TypeNet, as it retains the memory of previous typing activities, significantly enhancing its ability to discern and predict an individual's distinctive typing style, particularly in extensive texts. Engineered to be compatible with various input devices, both physical and touchscreen, TypeNet is adaptable across different platforms. It was rigorously evaluated using three distinct loss functions: softmax, contrastive, and triplet loss, which contributes to its versatility in application. For detailed architectural information, please refer to \cite{acien2021typenet}. Leveraging the capabilities of TypeNet, we adapted this model for our plagiarism detection initiatives, utilizing keystroke dynamics. Subsequent paragraphs will elaborate on the modifications made to the TypeNet architecture for this purpose. 

\textit{\textbf{System architecture:}} We enhanced the TypeNet architecture by integrating a Siamese network, which processes sequence pairs to differentiate between bona fide and assisted writing. The modifications to the architecture include: (1) Two LSTM layers, each producing a $128$-dimensional output, complemented by batch normalization layers and tanh activation functions. (2) A fully connected layer succeeding each LSTM, designed to map the embeddings to a uniform $128$-dimensional space. (3) Dropout layers added post each LSTM layer to mitigate overfitting, with a dropout rate of $0.5$ and a recurrent dropout of $0.2$. (4) The sequence length $M$ and batch size were varied during experiments, with $M$ ranging from $25$ to $500$ and batch sizes from $32$ to $512$. We explored different training durations, observing that performance typically stabilized between $50$ and $100$ epochs. The Adam optimizer was deployed with an adaptable learning rate ranging from $0.0001$ to $0.005$, and $L_2$ regularization was applied to prevent over-fitting further.

\textit{\textbf{Data Preprocessing:}} We standardized the sequence lengths to a predetermined length \(M\) by padding shorter sequences and clipping longer ones while adjusting \(M\) during hyperparameter tuning. The value of \(M\) ranged between $25$ and $500$, with batch sizes varying between $32$ and $512$, in powers of $2$. Sequences with excessive use of the Shift key (over $20$\%) and those significantly shorter than \(M\) (less than $50$\%) were excluded. Each keystroke sequence was characterized by three attributes: timestamps, keycodes, and key actions (KeyDown, KeyUp), normalized as follows: Timestamps were scaled to a range of $[0, 1]$ using min-max normalization. Keycodes were normalized by dividing each by $255$. Key actions were binary encoded, with KeyDown as $0$ and KeyUp as $1$. Siamese networks were trained using pairs of sequences from fixed and free-text contexts. Opposite sequence pairs were labeled $1$, and similar sequence pairs $0$. We maintained a balanced dataset to minimize biases.

\textit{\textbf{Loss Function:}} Contrary to the original TypeNet, which utilized Euclidean distance for comparisons, our model incorporates a modified loss function that employs cosine similarity to assess the closeness of embeddings from paired sequences. Training leverages Binary Cross Entropy Loss, with adjustments reflecting the cosine similarities. The decision threshold for classifying sequences as bona fide or assisted is determined by the Equal Error Rate (EER) point, derived from ROC analysis, which optimizes the balance between false acceptance and rejection rates. This threshold is then used to calculate the predicted output's Accuracy and $F_1$ score.
 
\subsection{Evaluation scenarios}
We assessed TypeNet's effectiveness as a plagiarism detector through various evaluation setups, focusing on the user, keyboard, context, and datasets. Each scenario utilized specific and agnostic modeling to measure the model’s adaptability and accuracy under diverse conditions.

We explored both \textbf{\textit{user-specific}} and \textbf{\textit{user-agnostic}} scenarios. In the \textbf{\textit{user-specific}} evaluation, we trained and tested the model independently with an $80$-$20$ split for each user’s data, ensuring no overlap between the training and testing datasets. We aggregated all possible user sequences from the datasets, then divided each sequence for training and testing to assess the model’s performance on sequences from users previously trained with different data from the same user. In the \textbf{\textit{user-agnostic}} approach, we assessed the model’s performance across various datasets to gauge its consistency and generalizability. This involved creating a divided environment by splitting the entire set of users into distinct training and testing groups, ensuring no user data overlapped between groups. We varied the ratios for training, validation, and testing, such as $50$-$25$-$25$ and $80$-$10$-$10$, to evaluate the model’s efficacy across diverse user groups.

In the \textbf{\textit{keyboard-specific}} setup, the model was trained and tested on data collected from the same keyboard type with an $80$-$20$ split, ensuring distinct sequences in each set. The \textbf{\textit{keyboard-agnostic}} evaluation involved training the model using data from three keyboard types and testing it on a fourth. This approach evaluated the model's ability to generalize across different keyboard types, capturing the subtleties of keystroke dynamics regardless of the keyboard used. For this evaluation, we employed the Buffalo dataset, which categorizes data based on keyboard type, differentiating between free and fixed sequences for detection purposes. For convenience, the \textit{Lenovo} keyboard is labeled as $K_{0}$, \textit{HP wireless} as $K_{1}$, \textit{Microsoft} as $K_{2}$, and the \textit{Apple Bluetooth} as $K_{3}$.
 
\textbf{\textit{Context-specific}} modeling entailed training and testing the model on homogeneous datasets, which maintained content consistency within each set. This approach assessed the model's effectiveness in managing data with similar linguistic contexts. In context-specific scenarios, we trained and tested the model on sequences from the same context, ensuring there was no overlap of sequences between the training and testing sets. An 80-20 split was implemented for sequences from each context for training and testing purposes. Conversely, \textbf{\textit{context-agnostic}} modeling presented a challenge to the model by training on datasets that covered two different topics and testing on a third, previously unseen topic. This method evaluated the model's ability to adapt and perform linguistic analysis across diverse contexts, assessing its proficiency with familiar and new content. For this evaluation, we utilized the SBU dataset, which is divided into three parts based on context: Gay Marriage, Gun Control, and Restaurant Feedback, labeled as \textit{GM}, \textit{GC}, and \textit{RF}, respectively.
 
In the \textbf{\textit{dataset-specific}} approach, we train and test the model using the same dataset, ensuring no sequence overlap between the training and testing sets. This method evaluates the model's performance within a consistent typing environment. Conversely, the \textbf{\textit{dataset-agnostic}} approach involves training the model on a combination of datasets and testing it on different datasets or combinations thereof. This strategy assesses the model’s ability to generalize and maintain effectiveness across diverse environments, which may differ significantly in their curation. By employing both evaluation methods, we aim to comprehensively understand the model's strengths and weaknesses across various data collection environments, ensuring its suitability and generalizability for real-world applications.

\subsection{Evaluation metrics}

We measured and reported Accuracy, $F_1$ score, False Acceptance Rate (FAR), and False Rejection Rate (FRR) for each detector and scenario.

\textbf{\textit{Accuracy}} assesses the effectiveness of the plagiarism detection system by calculating the ratio of correctly identified documents (both true positives and true negatives) to the total number of evaluated documents. High Accuracy indicates that the system effectively distinguishes between assisted (true positives) and bona fide (true negatives) submissions, ensuring fair student evaluations.

\textbf{\textit{$F_1$ score}} represents the harmonic mean of precision and recall, providing a balanced measure of the system’s precision and ability to identify all actual instances of plagiarism. A strong $F_1$ score implies that the system reliably identifies a high percentage of true plagiarism cases while maintaining a low rate of false positives. This balance is crucial to prevent wrongful accusations against students, which could damage their academic reputation and trust in the educational system.

\textbf{\textit{False Acceptance Rate (FAR)}} reflects the likelihood that the system will fail to detect actual plagiarism, incorrectly labeling a plagiarized document as bona fide. It is essential to reduce FAR to ensure that all students are assessed fairly and that genuine efforts are acknowledged properly.

\textbf{\textit{False Rejection Rate (FRR)}} measures the probability that the system will mistakenly classify a bona fide document as plagiarized. Maintaining a low FRR is vital to protect students from unjust accusations.

\section{Results and discussion} 
\label{ResultsDiscussion}
We present the performance of each model across all evaluation scenarios in Tables \ref{tab:my-table1} and \ref{tab:my-table2}, which are further explained in the subsequent paragraphs.

\begin{table}[htp]
\centering
\caption{The table summarizes the performance of the TypeNet architecture across various datasets and scenarios, highlighting metrics such as Accuracy, $F_1$ scores, FAR, and FRR. It encompasses keyboard-specific scenarios ($K_0, K_1, K_2, K_3$ represent different keyboards), context-specific scenarios (GM, GC, RF), user-specific merged data from all datasets, and dataset-specific evaluations (SBU - S, Proposed - P, Buffalo - B). Notably, the lowest observed Accuracy is 74.98\% in the keyboard-specific $K_3$ scenario, whereas the highest reaches 85.72\% in the dataset-specific Buffalo scenario. $F_1$ scores span from a low of 73.34\% ($K_3$) to a high of 84.72\% (B). The merged user-specific data demonstrates robust performance, with both Accuracy and $F_1$ score exceeding 81\%. Among dataset-specific evaluations, the Buffalo dataset stands out as the top performer, achieving an Accuracy of 85.72\% and an $F_1$ score of 84.72\%.}
\vspace{0.15in}
\begin{tabular}{|cccccc|}
\hline
\multicolumn{2}{|c|}{\textbf{Dataset}}                                    & \multicolumn{4}{c|}{\textbf{TypeNet}}                                                                                    \\ \hline
\multicolumn{1}{|c|}{\textbf{Train}} & \multicolumn{1}{c|}{\textbf{Test}} & \multicolumn{1}{c|}{\textbf{Acc.}} & \multicolumn{1}{c|}{\textbf{$F_1$}} & \multicolumn{1}{c|}{\textbf{FAR}} & \textbf{FRR} \\ \hline
\multicolumn{6}{|c|}{\textit{Keyboard Specific}}                                                                                                                                                     \\ \hline
\multicolumn{1}{|c|}{$K_0$}             & \multicolumn{1}{c|}{$K_0$}            & \multicolumn{1}{c|}{84.64}         & \multicolumn{1}{c|}{83.45}       & \multicolumn{1}{c|}{25.38}        & 19.83        \\ \hline
\multicolumn{1}{|c|}{$K_1$}             & \multicolumn{1}{c|}{$K_1$}            & \multicolumn{1}{c|}{80.02}         & \multicolumn{1}{c|}{78.91}       & \multicolumn{1}{c|}{29.11}        & 24.19        \\ \hline
\multicolumn{1}{|c|}{$K_2$}             & \multicolumn{1}{c|}{$K_2$}            & \multicolumn{1}{c|}{77.77}         & \multicolumn{1}{c|}{76.58}       & \multicolumn{1}{c|}{32.14}        & 25.12        \\ \hline
\multicolumn{1}{|c|}{$K_3$}             & \multicolumn{1}{c|}{$K_3$}            & \multicolumn{1}{c|}{74.98}         & \multicolumn{1}{c|}{73.34}       & \multicolumn{1}{c|}{32.6}         & 30.16        \\ \hline
\multicolumn{6}{|c|}{\textit{Context Specific}}                                                                                                                                                      \\ \hline
\multicolumn{1}{|c|}{GM}             & \multicolumn{1}{c|}{GM}            & \multicolumn{1}{c|}{80.24}         & \multicolumn{1}{c|}{81.52}       & \multicolumn{1}{c|}{30.81}        & 21.27        \\ \hline
\multicolumn{1}{|c|}{GC}             & \multicolumn{1}{c|}{GC}            & \multicolumn{1}{c|}{79.39}         & \multicolumn{1}{c|}{80.67}       & \multicolumn{1}{c|}{34.62}        & 22.33        \\ \hline
\multicolumn{1}{|c|}{RF}             & \multicolumn{1}{c|}{RF}            & \multicolumn{1}{c|}{76.52}         & \multicolumn{1}{c|}{78.01}       & \multicolumn{1}{c|}{34.02}        & 31.13        \\ \hline
\multicolumn{6}{|c|}{\textit{User Specific}}                                                                                                                                                         \\ \hline
\multicolumn{1}{|c|}{Merged}       & \multicolumn{1}{c|}{Merged}      & \multicolumn{1}{c|}{81.86}         & \multicolumn{1}{c|}{81.85}       & \multicolumn{1}{c|}{23.71}        & 26.24        \\ \hline
\multicolumn{6}{|c|}{\textit{Dataset Specific}}                                                                                                                                                      \\ \hline
\multicolumn{1}{|c|}{S}            & \multicolumn{1}{c|}{S}           & \multicolumn{1}{c|}{80.85}         & \multicolumn{1}{c|}{83.31}       & \multicolumn{1}{c|}{18.83}        & 32.93        \\ \hline
\multicolumn{1}{|c|}{P}       & \multicolumn{1}{c|}{P}      & \multicolumn{1}{c|}{81.04}         & \multicolumn{1}{c|}{82.16}       & \multicolumn{1}{c|}{22.6}         & 28.2         \\ \hline
\multicolumn{1}{|c|}{B}        & \multicolumn{1}{c|}{B}       & \multicolumn{1}{c|}{85.72}         & \multicolumn{1}{c|}{84.72}       & \multicolumn{1}{c|}{17.64}        & 23.91        \\ \hline
\end{tabular}
\label{tab:my-table1}
\end{table}
 
The \textit{\textbf{user-specific}} models, where the training and testing sets include keystroke sequences from the same user with no overlap, achieved Accuracy and $F_1$ scores of $81.86\%$ and $81.85\%$. FAR and FRR were recorded at $23.71\%$ and $26.24\%$, respectively, as illustrated in Table \ref{tab:my-table1}. 

Conversely, \textbf{\textit{user-agnostic}} models achieved an accuracy ranging from $63.56\%$ to $66.54\%$. The FAR and FRR ranged from $38.82\%$ to $42.51\%$ and $39.57\%$ to $39.86\%$, respectively (Table \ref{tab:my-table2}).

These findings indicate that the model performs better with keystroke sequences from the same user than those from different users, demonstrating limited generalizability across user keystroke patterns. This is consistent with the significant variation in typing behavior among users, even when typing identical content-- offering insights into the effectiveness of continuous authentication through keystroke dynamics.

In \textit{\textbf{keyboard-specific}} scenarios, where the keyboard type is consistent and known, the models achieve an accuracy in the range of $74.98\%$ to $84.64\%$, and $F_1$ scores range from $73.34\%$ to $83.45\%$. 

In \textbf{\textit{keyboard-agnostic}} scenarios, the models attained an accuracy between $78.11\%$ and $80.54\%$, and $F_1$ scores between $76.89\%$ and $78.77\%$. In both cases, the FAR and FRR remain below $30\%$. The minimal differences in accuracy, $F_1$ scores, and error rates suggest that the proposed plagiarism detector is robust regardless of the type of keyboard used.

In \textbf{\textit{context-specific}} scenarios (see Table \ref{tab:my-table1}), TypeNet achieved an Accuracy in the range of $76.52\%$ to $80.24\%$, $F_1$ scores from $78.01\%$ to $81.52\%$, FAR from $30.81\%$ to $34.62\%$, and FRR from $21.27\%$ to $31.13\%$. These metrics illustrate the model's effectiveness in environments that align with its training conditions. However, performance slightly drops in \textbf{\textit{context-agnostic}} scenarios (see Table \ref{tab:my-table2}). Accuracy ranged from $70.21\%$ to $78.67\%$. $F_1$ scores were between $70.23\%$ and $78.24\%$. FAR increased, ranging from $32.30\%$ to $39.65\%$, and FRR extended from $24.10\%$ to $34.73\%$. The performance remains strong despite these reductions, indicating only modest declines from the context-specific results. This underscores TypeNet's adaptability across varying contexts.
 
\begin{table}[htp]
\centering
\caption{The performance of the TypeNet architecture across various platform-agnostic scenarios is summarized in the table, with metrics including Accuracy, $F_1$ scores, FAR, and FRR. The Merged data consists of user sequences combined from all datasets, where "S" denotes the SBU dataset, "P" stands for the Proposed dataset, and "B" for the Buffalo dataset. The architecture achieves its best and most consistent performance in keyboard-agnostic scenarios, with Accuracy ranging from $78.11\%$ to $80.54\%$ and robust $F_1$ scores. Context-agnostic scenarios exhibit variability, with the lowest Accuracy recorded at $70.21\%$. In user-agnostic scenarios, performance notably drops below $70\%$ in certain configurations, reflecting the model's challenges in adapting to diverse user behaviors. Dataset-agnostic testing uncovers the architecture's difficulties in generalizing across unfamiliar data, leading to significant performance reductions, especially in configurations where training and testing datasets do not overlap, yielding the lowest accuracies and highest error rates.}
\vspace{.15in}
\label{tab:my-table2}
\begin{tabular}{|cccccc|}
\hline
\multicolumn{2}{|c|}{\textbf{Dataset}}                                    & \multicolumn{4}{c|}{\textbf{TypeNet}}                                                                                    \\ \hline
\multicolumn{1}{|c|}{\textbf{Train}} & \multicolumn{1}{c|}{\textbf{Test}} & \multicolumn{1}{c|}{\textbf{Acc.}} & \multicolumn{1}{c|}{\textbf{$F_1$}} & \multicolumn{1}{c|}{\textbf{FAR}} & \textbf{FRR} \\ \hline
\multicolumn{6}{|c|}{\textit{Keyboard Agnostic}}                                                                                                                                                     \\ \hline
\multicolumn{1}{|c|}{$K_{0,1,2}$}         & \multicolumn{1}{c|}{$K_3$}            & \multicolumn{1}{c|}{78.11}         & \multicolumn{1}{c|}{75.88}       & \multicolumn{1}{c|}{28.04}        & 29.01        \\ \hline
\multicolumn{1}{|c|}{$K_{0,1,3}$}         & \multicolumn{1}{c|}{$K_2$}            & \multicolumn{1}{c|}{79.71}         & \multicolumn{1}{c|}{78.06}       & \multicolumn{1}{c|}{27.5}         & 28.37        \\ \hline
\multicolumn{1}{|c|}{$K_{0,2,3}$}         & \multicolumn{1}{c|}{$K_1$}            & \multicolumn{1}{c|}{80.54}         & \multicolumn{1}{c|}{78.77}       & \multicolumn{1}{c|}{24.01}        & 28.37        \\ \hline
\multicolumn{1}{|c|}{$K_{1,2,3}$}         & \multicolumn{1}{c|}{$K_0$}            & \multicolumn{1}{c|}{78.96}         & \multicolumn{1}{c|}{76.89}       & \multicolumn{1}{c|}{29.15}        & 24.95        \\ \hline
\multicolumn{6}{|c|}{\textit{Context Agnostic}}                                                                                                                                                      \\ \hline
\multicolumn{1}{|c|}{(GM, RF)}        & \multicolumn{1}{c|}{GC}            & \multicolumn{1}{c|}{72.96}         & \multicolumn{1}{c|}{72.4}        & \multicolumn{1}{c|}{38.3}         & 28.56        \\ \hline
\multicolumn{1}{|c|}{(GC, RF)}        & \multicolumn{1}{c|}{GM}            & \multicolumn{1}{c|}{78.67}         & \multicolumn{1}{c|}{78.24}       & \multicolumn{1}{c|}{32.3}         & 24.1         \\ \hline
\multicolumn{1}{|c|}{(GM, GC)}        & \multicolumn{1}{c|}{RF}            & \multicolumn{1}{c|}{70.21}         & \multicolumn{1}{c|}{70.23}       & \multicolumn{1}{c|}{39.65}        & 34.73        \\ \hline
\multicolumn{6}{|c|}{\textit{User Agnostic}}                                                                                                                                                         \\ \hline
\multicolumn{2}{|c|}{Merged (80-10-10)}                                 & \multicolumn{1}{c|}{63.56}         & \multicolumn{1}{c|}{62.98}       & \multicolumn{1}{c|}{42.51}        & 39.57        \\ \hline
\multicolumn{2}{|c|}{Merged (50-25-25)}                                 & \multicolumn{1}{c|}{66.54}         & \multicolumn{1}{c|}{66.22}       & \multicolumn{1}{c|}{38.82}        & 39.86        \\ \hline
\multicolumn{6}{|c|}{\textit{Dataset Agnostic}}                                                                                                                                                      \\ \hline
\multicolumn{1}{|c|}{(S, P)}           & \multicolumn{1}{c|}{B}             & \multicolumn{1}{c|}{68.72}         & \multicolumn{1}{c|}{66.21}       & \multicolumn{1}{c|}{33.13}        & 39.66        \\ \hline
\multicolumn{1}{|c|}{(S, B)}           & \multicolumn{1}{c|}{P}             & \multicolumn{1}{c|}{73.23}         & \multicolumn{1}{c|}{72.65}       & \multicolumn{1}{c|}{27.95}        & 40.22        \\ \hline
\multicolumn{1}{|c|}{(P, B)}           & \multicolumn{1}{c|}{S}             & \multicolumn{1}{c|}{52.24}         & \multicolumn{1}{c|}{61.86}       & \multicolumn{1}{c|}{47.53}        & 48.51        \\ \hline
\multicolumn{1}{|c|}{S}              & \multicolumn{1}{c|}{(P, B)}          & \multicolumn{1}{c|}{59.73}         & \multicolumn{1}{c|}{57.03}       & \multicolumn{1}{c|}{41.36}        & 44.29        \\ \hline
\multicolumn{1}{|c|}{P}              & \multicolumn{1}{c|}{(S, B)}          & \multicolumn{1}{c|}{56.17}         & \multicolumn{1}{c|}{54.13}       & \multicolumn{1}{c|}{44.95}        & 45.72        \\ \hline
\multicolumn{1}{|c|}{B}              & \multicolumn{1}{c|}{(S, P)}          & \multicolumn{1}{c|}{53.57}         & \multicolumn{1}{c|}{53.16}       & \multicolumn{1}{c|}{46.32}        & 48.01        \\ \hline
\end{tabular}
\end{table}

In \textbf{\textit{dataset-specific}} scenarios, the TypeNet model demonstrates significant improvement, with each case achieving an accuracy and $F_1$ score above $80\%$, and the highest accuracy recorded at $85.72\%$. FAR and FRR were observed below $23\%$ and $30\%$, with the lowest values at $17.64\%$ and $23.91\%$, respectively (see Table \ref{tab:my-table2}). This indicates strong performance when the model is trained and tested on datasets curated for a particular subject matter. 

In \textbf{\textit{dataset-agnostic}} scenarios, where the model is trained across multiple datasets, we observed two instances with higher Accuracy and lower FAR/FRR when training incorporated a mix of keystroke sequences from the SBU dataset (S) and either the Buffalo (B) or Proposed (P) datasets. This approach maximizes the training data pool, resulting in improved model performance. The model performs best when trained on a combination of the SBU and Buffalo datasets and tested on the Proposed dataset, achieving an accuracy of $73.23\%$, an $F_1$ score of $72.65\%$, FAR of $27.95\%$, and FRR of $40.22\%$. However, in cases where the training data is less diverse, both the accuracy and $F_1$ score fall below $60\%$, with FAR and FRR exceeding $40\%$ and $45\%$, respectively, as illustrated in Table \ref{tab:my-table2}. These findings suggest that TypeNet requires more comprehensive data to generalize effectively across different environments. However, with a larger dataset, the model's performance significantly improves, highlighting the importance of extensive training data to enhance its robustness.

\subsection{Discussion}
The proposed TypeNet-based plagiarism detector exhibited varied error rates across different data collection scenarios and contexts. While achieving promising performances in scenario-specific settings, the detector struggled to generalize across scenarios. This aligns with findings from previous keystroke studies, including those involving TypeNet \cite{acien2021typenet, bours2018cross}. Further research is needed to develop a model generalizing across various datasets and collection environments.

A recent study indicates that simple statistical models outperform advanced deep learning models like TypeNet in keystroke classification \cite{wahab2023simple}. Consequently, we plan to explore the effectiveness of latency and word-level features and classifiers such as Instance-based Tail Area Density (ITAD), SVM, kNN, and Random Forest \cite{ayotte_itad, kamalov2021machine}. Additionally, we intend to incorporate strategies used in recent AI vs. Human detectors \cite{pu2023chatgpt, dugan2023real, Herbold2023, CASAL2023100068, BAiVsHuman, AiVsHumanCS1, AiVsHuman}, alongside content-independent linguistic features (such as intensifiers, hedging, politeness markers, negations, filler words, function words), and leverage transformer-based architectures \cite{vaswani2017attention} in future developments. Access to an individual’s keystroke data beforehand will allow us to create personalized models to enhance the system’s accuracy and adaptability.

While we have reported on the performance of the TypeNet architecture, a recent study \cite{wahab2023simple} prompts us to evaluate traditional keystroke modeling approaches in future research. We simplified our experimental setup by assuming copy-paste actions would be disabled using tools such as lockdown browsers during examinations. However, this assumption may not hold in all examination settings, as some may allow copy-pasting within the test editor. Monitoring these actions and clipboard data could enhance our method's performance.

Moreover, a detailed theoretical analysis of the distinguishability of typing patterns would provide further insights to our proposed method. Nevertheless, existing research \cite{trezise2019detecting} indicates that bona fide (free) writing typically involves shorter bursts of typing, longer pauses, and more frequent revisions compared to assisted (transcription-like) tasks. These observations underscore the necessity for further investigation into typing pattern variability.
 
\section{Limitations}
\label{limitations}
While this study underscores the potential of keystroke dynamics for detecting academic dishonesty within the context of large language models (LLMs), several limitations necessitate further investigation. Firstly, the findings presented are based on a limited number of users and datasets created in controlled environments. These datasets may not adequately represent the diverse typing behaviors of a broader population, thereby impacting the model's accuracy and applicability across various demographic groups and educational settings, such as creative writing, timed responses, or coding tasks. Moreover, a high rate of false positives could lead to wrongful accusations and subsequent distrust in the detection system, highlighting the need for more sophisticated models to reduce errors.

Additionally, the method's effectiveness varied significantly with external factors such as keyboard layout, academic contexts, and individual typing habits. Future research should focus on developing more robust models that can adapt to these variables, ensuring consistent performance across different hardware configurations and user backgrounds.

\section{Conclusions and future work}
\label{ConclusionFutureWorks}
We explored the potential of keystroke dynamics to detect instances of academic dishonesty facilitated by advanced generative AI tools. Based on keystroke dynamics, our model effectively distinguished between bona fide and AI-assisted writings at the keystroke level, showing promising performance across various testing scenarios. These results underscore the value of keystroke dynamics as a reliable supplement to traditional plagiarism detection tools, particularly crucial in an era dominated by large language models (LLMs) capable of generating human-like texts.

The study also highlights the operational challenges of implementing keystroke dynamics across diverse contexts. Although the model demonstrated reasonable effectiveness under controlled experimental conditions, applying it in real-world settings necessitates further extensive research. Moving forward, we aim to broaden our research to encompass a wider array of users, educational settings, and forms of academic dishonesty, including AI assistance, integration of internet content, and copy-paste-edit behaviors. Additionally, we plan to refine our detection algorithms to improve their accuracy and effectiveness, ensuring they are adaptable across a diverse range of user demographics and academic contexts.

\section{Acknowledgment} We thank the participants and anonymous reviewers for their invaluable contributions to this study. Rajiv Ratn Shah was partly supported by the Infosys Center for Artificial Intelligence, the Center for Design and New Media, and the Center of Excellence in Healthcare at IIIT Delhi, India.

{\small
\bibliography{egbib}

\begin{thebibliography}{10}

\bibitem{rao2008plagiarism}
KR~Rao.
\newblock Plagiarism, a scourge.
\newblock {\em Current Science}, 2008.

\bibitem{anderson2009avoiding}
Irene Anderson.
\newblock Avoiding plagiarism in academic writing.
\newblock {\em Nursing standard}, 2009.

\bibitem{pecorari2008academic}
Diane Pecorari.
\newblock Academic writing and plagiarism.
\newblock {\em Bloomsbury Publishing}, 2008.

\bibitem{roig2015avoiding}
Miguel Roig.
\newblock Avoiding plagiarism, self-plagiarism, and other questionable writing practices: A guide to ethical writing.
\newblock {\em The Office of Research Integrity (ORI)}, 2015.

\bibitem{hoq2024detecting}
Muntasir Hoq, Yang Shi, Juho Leinonen, Damilola Babalola, Collin Lynch, Thomas Price, and Bita Akram.
\newblock Detecting chatgpt-generated code submissions in a cs1 course using machine learning models.
\newblock {\em SIGCSE 2024}, 2024.

\bibitem{zhang2011coarse}
Haijun Zhang and Tommy~WS Chow.
\newblock A coarse-to-fine framework to efficiently thwart plagiarism.
\newblock {\em Pattern Recognition}, 2011.

\bibitem{pu2023chatgpt}
Dongqi Pu and Vera Demberg.
\newblock Chatgpt vs. human-authored text: Insights into controllable text summarization and sentence style transfer.
\newblock {\em arXiv preprint arXiv:2306.07799}, 2023.

\bibitem{dugan2023real}
Liam Dugan, Daphne Ippolito, Arun Kirubarajan, Sherry Shi, and Chris Callison-Burch.
\newblock Real or fake text?: Investigating human ability to detect boundaries between human-written and machine-generated text.
\newblock In {\em Proceedings of the AAAI Conference on Artificial Intelligence}, 2023.

\bibitem{Herbold2023}
Steffen Herbold, Annette Hautli-Janisz, Ute Heuer, Zlata Kikteva, and Alexander Trautsch.
\newblock A large-scale comparison of human-written versus chatgpt-generated essays.
\newblock {\em Scientific Reports}, 2023.

\bibitem{CASAL2023100068}
J.~Elliott Casal and Matt Kessler.
\newblock Can linguists distinguish between chatgpt/ai and human writing?: A study of research ethics and academic publishing.
\newblock {\em Research Methods in Applied Linguistics}, 2023.

\bibitem{BAiVsHuman}
Tony {Berber Sardinha}.
\newblock Ai-generated vs human-authored texts: A multidimensional comparison.
\newblock {\em Applied Corpus Linguistics}, 2024.

\bibitem{AiVsHumanCS1}
Hosam Alamleh, Ali Abdullah~S. AlQahtani, and AbdElRahman ElSaid.
\newblock Distinguishing human-written and chatgpt-generated text using machine learning.
\newblock In {\em 2023 Systems and Information Engineering Design Symposium (SIEDS)}, 2023.

\bibitem{AiVsHuman}
Niful {Islam}, Debopom {Sutradhar}, Humaira {Noor}, Jarin~Tasnim {Raya}, Monowara {Tabassum Maisha}, and Dewan~Md {Farid}.
\newblock {Distinguishing Human Generated Text From ChatGPT Generated Text Using Machine Learning}.
\newblock {\em arXiv e-prints}, 2023.

\bibitem{khurana2023synthesizing}
Varun Khurana, Yaman Kumar, Nora Hollenstein, Rajesh Kumar, and Balaji Krishnamurthy.
\newblock Synthesizing human gaze feedback for improved {NLP} performance.
\newblock In {\em Proceedings of the 17th Conference of the European Chapter of the Association for Computational Linguistics}, 2023.

\bibitem{BehavioralNLP2}
Paul Prasse, David Reich, Silvia Makowski, Tobias Scheffer, and Lena Jäger.
\newblock Improving cognitive-state analysis from eye gaze with synthetic eye-movement data.
\newblock {\em Computers and Graphics}, 2024.

\bibitem{MONROSE2000351}
Fabian Monrose and Aviel~D. Rubin.
\newblock Keystroke dynamics as a biometric for authentication.
\newblock {\em Future Generation Computer Systems}, 2000.

\bibitem{Joyce1990}
Rick Joyce and Gopal Gupta.
\newblock Identity authentication based on keystroke latencies.
\newblock {\em Commun. ACM}, 1990.

\bibitem{62613}
S.~Bleha, C.~Slivinsky, and B.~Hussien.
\newblock Computer-access security systems using keystroke dynamics.
\newblock {\em IEEE Transactions on Pattern Analysis and Machine Intelligence}, 1990.

\bibitem{press1980authentication}
S~Press.
\newblock Authentication by keystroke timing: Some preliminary results.
\newblock {\em Rand Report}, 1980.

\bibitem{teh2013survey}
Pin~Shen Teh, Andrew Beng~Jin Teoh, Shigang Yue, et~al.
\newblock A survey of keystroke dynamics biometrics.
\newblock {\em The Scientific World Journal}, 2013.

\bibitem{TypingPhoneBTAS2016}
Rajesh Kumar, Vir~V. Phoha, and Abdul Serwadda.
\newblock Continuous authentication of smartphone users by fusing typing, swiping, and phone movement patterns.
\newblock In {\em 2016 IEEE 8th International Conference on Biometrics Theory, Applications, and Systems (BTAS)}, 2016.

\bibitem{alzahrani2011understanding}
Salha~M Alzahrani, Naomie Salim, and Ajith Abraham.
\newblock Understanding plagiarism linguistic patterns, textual features, and detection methods.
\newblock {\em IEEE Transactions on Systems, Man, and Cybernetics, Part C (Applications and Reviews)}, 2011.

\bibitem{deane2015exploring}
Paul Deane and Mo~Zhang.
\newblock Exploring the feasibility of using writing process features to assess text production skills.
\newblock {\em ETS Research Report Series}, 2015.

\bibitem{agarwal-nancy-contract}
Nancy Agarwal, Nils~Folvik Danielsen, Per~Kristian Gravdal, and Patrick Bours.
\newblock Contract cheat detection using biometric keystroke dynamics.
\newblock In {\em 2022 20th International Conference on Emerging eLearning Technologies and Applications (ICETA)}, 2022.

\bibitem{kochegurova2022hidden}
Elena~A Kochegurova and Roman~P Zateev.
\newblock Hidden monitoring based on keystroke dynamics in online examination system.
\newblock {\em Programming and Computer Software}, 2022.

\bibitem{mungai2017using}
Peter~Kimani Mungai and Runhe Huang.
\newblock Using keystroke dynamics in a multi-level architecture to protect online examinations from impersonation.
\newblock In {\em 2017 IEEE 2nd International Conference on Big Data Analysis (ICBDA)}. IEEE, 2017.

\bibitem{hayes2012modeling}
John~R Hayes.
\newblock Modeling and remodeling writing.
\newblock {\em Written communication}, 2012.

\bibitem{flower1981cognitive}
Linda Flower and John~R Hayes.
\newblock A cognitive process theory of writing.
\newblock {\em College composition and communication}, 1981.

\bibitem{balagani2013investigating}
Kiran~S Balagani and NEW YORK INST OF TECH~OLD WESTBURY.
\newblock Investigating cognitive rhythms as a new modality for continuous authentication.
\newblock Technical report, Tech. Rep. DTIC Document, 2013.

\bibitem{zhang2021using}
Mo~Zhang, Hongwen Guo, and Xiang Liu.
\newblock Using keystroke analytics to understand cognitive processes during writing.
\newblock {\em International Educational Data Mining Society}, 2021.

\bibitem{lim2015using}
Yee~Mei Lim, Aladdin Ayesh, and Martin Stacey.
\newblock Using mouse and keyboard dynamics to detect cognitive stress during mental arithmetic.
\newblock In {\em Intelligent Systems in Science and Information 2014: Extended and Selected Results from the Science and Information Conference 2014}, 2015.

\bibitem{patel2011evaluation}
Ahmed Patel, Kaveh Bakhtiyari, and Mona Taghavi.
\newblock Evaluation of cheating detection methods in academic writings.
\newblock {\em Library Hi Tech}, 2011.

\bibitem{kamalov2021machine}
Firuz Kamalov, Hana Sulieman, and David Santandreu~Calonge.
\newblock Machine learning-based approach to exam cheating detection.
\newblock {\em Plos one}, 2021.

\bibitem{roa2022automated}
M~Roa’a, Ibtisam~A Aljazaery, and AHM Alaidi.
\newblock Automated cheating detection based on video surveillance in the examination classes.
\newblock {\em International Journal of Interactive Mobile Technologies}, 2022.

\bibitem{tiong2021cheating}
Leslie Ching~Ow Tiong and HeeJeong~Jasmine Lee.
\newblock E-cheating prevention measures: detection of cheating at online examinations using deep learning approach--a case study.
\newblock {\em arXiv preprint arXiv:2101.09841}, 2021.

\bibitem{ayotte2020fast}
Blaine Ayotte, Mahesh Banavar, Daqing Hou, and Stephanie Schuckers.
\newblock Fast free-text authentication via instance-based keystroke dynamics.
\newblock {\em IEEE Transactions on Biometrics, Behavior, and Identity Science}, 2020.

\bibitem{monrose1999password}
Fabian Monrose, Michael~K Reiter, and Susanne Wetzel.
\newblock Password hardening based on keystroke dynamics.
\newblock In {\em Proceedings of the 6th ACM Conference on Computer and Communications Security}, 1999.

\bibitem{KeystrokeSoftBiometrics}
Vishaal Udandarao, Mohit Agrawal, Rajesh Kumar, and Rajiv~Ratn Shah.
\newblock On the inference of soft biometrics from typing patterns collected in a multi-device environment.
\newblock In {\em 2020 IEEE Sixth International Conference on Multimedia Big Data (BigMM)}, 2020.

\bibitem{Monaro2018Covert}
Mirella Monaro, Claudia Galante, Riccardo Spolaor, et~al.
\newblock Covert lie detection using keyboard dynamics.
\newblock {\em Scientific Reports}, 2018.

\bibitem{banerjee2014_emnlp}
Ritwik Banerjee, Song Feng, Jun~Seok Kang, and Yejin Choi.
\newblock Keystroke patterns as prosody in digital writings: A case study with deceptive reviews and essays.
\newblock In {\em Proceedings of the 2014 Conference on Empirical Methods in Natural Language Processing (EMNLP)}, Doha, Qatar, 2014. Association for Computational Linguistics.

\bibitem{morales2020keystroke}
Aythami Morales, Alejandro Acien, Julian Fierrez, John~V. Monaco, Ruben Tolosana, Ruben Vera, and Javier Ortega-Garcia.
\newblock Keystroke biometrics in response to fake news propagation in a global pandemic.
\newblock In {\em 2020 IEEE 44th Annual Computers, Software, and Applications Conference (COMPSAC)}, 2020.

\bibitem{kuruvilla2024spotting}
Alvin Kuruvilla, Rojanaye Daley, and Rajesh Kumar.
\newblock Spotting fake profiles in social networks via keystroke dynamics.
\newblock In {\em 2024 IEEE 21st Consumer Communications and Networking Conference (CCNC)}, 2024.

\bibitem{tsimperidis2018r}
Ioannis Tsimperidis, Paul~D Yoo, Kamal Taha, Alexios Mylonas, and Vasilis Katos.
\newblock R 2 bn: An adaptive model for keystroke-dynamics-based educational level classification.
\newblock {\em IEEE Transactions on Cybernetics}, 2018.

\bibitem{plank2018predicting}
Barbara Plank.
\newblock Predicting authorship and author traits from keystroke dynamics.
\newblock In {\em Proceedings of the Second Workshop on Computational Modeling of People’s Opinions, Personality, and Emotions in Social Media}, 2018.

\bibitem{email-author-goodman}
Robert Goodman, Matthew Hahn, Madhuri Marella, Christina Ojar, and Sandy Westcott.
\newblock The use of stylometry for email author identification: A feasibility study.
\newblock {\em Proc. Student/Faculty Research Day}, 2007.

\bibitem{7823894}
Yan Sun, Hayreddin Ceker, and Shambhu Upadhyaya.
\newblock Shared keystroke dataset for continuous authentication.
\newblock In {\em 2016 IEEE International Workshop on Information Forensics and Security (WIFS)}, 2016.

\bibitem{lettner2010questions}
Heather Lettner-Rust.
\newblock Questions of transfer: Writers' perspective on familiar/unfamiliar writing tasks in a capstone writing course.
\newblock {\em Psychology}, 2010.

\bibitem{trezise2019detecting}
Kelly Trezise, Tracii Ryan, Paula de~Barba, and Gregor Kennedy.
\newblock Detecting academic misconduct using learning analytics.
\newblock {\em Journal of Learning Analytics}, 2019.

\bibitem{acien2021typenet}
Alejandro Acien, Aythami Morales, John~V. Monaco, Ruben Vera-Rodriguez, and Julian Fierrez.
\newblock Typenet: Deep learning keystroke biometrics.
\newblock {\em IEEE Transactions on Biometrics, Behavior, and Identity Science}, 2022.

\bibitem{bours2018cross}
Patrick Bours and J{\O}rgen Ellingsen.
\newblock Cross keyboard keystroke dynamics.
\newblock In {\em 2018 1st International Conference on Computer Applications and Information Security (ICCAIS)}. IEEE, 2018.

\bibitem{wahab2023simple}
Ahmed Wahab and Daqing Hou.
\newblock When simple statistical algorithms outperform deep learning: A case of keystroke dynamics [when simple statistical algorithms outperform deep learning: A case of keystroke dynamics].
\newblock In {\em Proceedings of the 12th International Conference on Pattern Recognition Applications and Methods ICPRAM}, 2023.

\bibitem{ayotte_itad}
Blaine Ayotte, Mahesh Banavar, Daqing Hou, and Stephanie Schuckers.
\newblock Fast free-text authentication via instance-based keystroke dynamics.
\newblock {\em IEEE Transactions on Biometrics, Behavior, and Identity Science}, 2020.

\bibitem{vaswani2017attention}
Ashish Vaswani, Noam Shazeer, Niki Parmar, Jakob Uszkoreit, Llion Jones, Aidan~N. Gomez, {\L}ukasz Kaiser, and Illia Polosukhin.
\newblock Attention is all you need.
\newblock {\em Advances in neural information processing systems}, 2017.

\end{thebibliography}
}

\end{document}